\documentclass[10pt,twocolumn,letterpaper]{article}

\usepackage{icb}
\usepackage{times}
\usepackage{epsfig}
\usepackage{graphicx}
\usepackage{amsmath}
\usepackage{amssymb}
\usepackage{multirow}
\usepackage{textcomp}
\usepackage{colortbl}
\usepackage[table]{xcolor}
\usepackage{booktabs}
\usepackage{url}

\usepackage{subcaption}

\definecolor{myblue}{rgb}{0,0,1}
\definecolor{myred}{rgb}{0.8, 0, 0}
\definecolor{mygreen}{rgb}{0, 0.6, 0}

\newcommand{\cellred}[1]{\cellcolor{red!45} {#1}}
\newcommand{\cellgreen}[1]{\cellcolor{green!45} {#1}}

\icbfinalcopy 

\ificbfinal\fi

\begin{document}

\title{Iris Recognition with Image Segmentation Employing Retrained Off-the-Shelf Deep Neural Networks}


\author{Daniel Kerrigan\\
University of Notre Dame\\
Notre Dame, Indiana, USA\\
{\tt\small dkerriga@alumni.nd.edu}
\and
Mateusz Trokielewicz\\
Research and Academic Computer Network\\
Warsaw, Poland\\
{\tt\small mateusz.trokielewicz@nask.pl}\\
\and
Adam Czajka\\
University of Notre Dame\\
Notre Dame, Indiana, USA\\
{\tt\small aczajka@nd.edu}\\
\and
Kevin W. Bowyer\\
University of Notre Dame\\
Notre Dame, Indiana, USA\\
{\tt\small kwb@nd.edu}\\
}

\maketitle
\thispagestyle{empty}

\begin{figure*}[!htb]
    \centering
    \begin{subfigure}[t]{0.19\textwidth}
        \includegraphics[width=\textwidth]{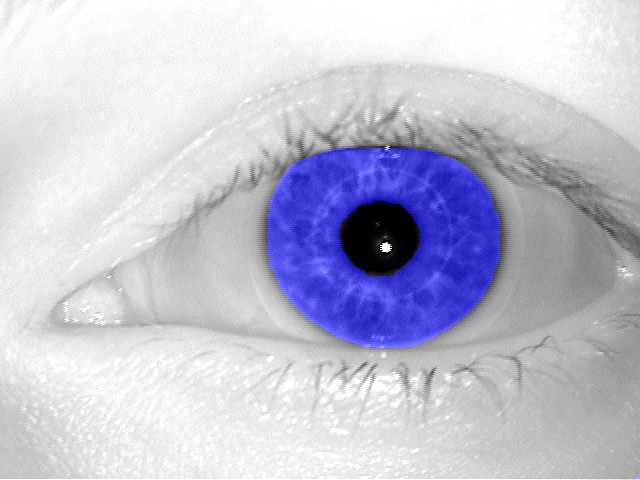}\\
        \includegraphics[width=\textwidth]{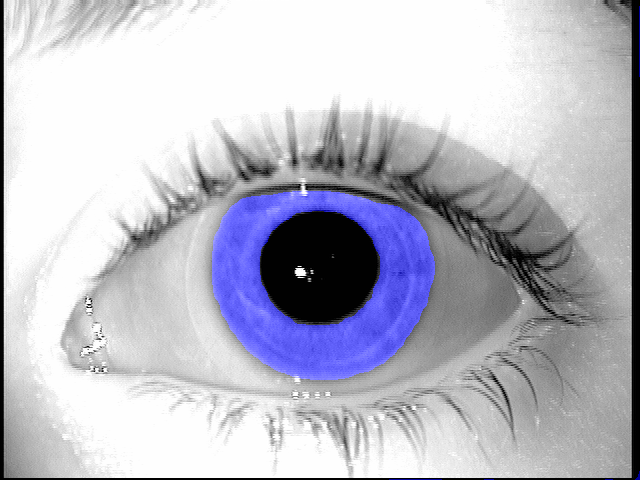}
        \caption{}
    \end{subfigure}
    \begin{subfigure}[t]{0.19\textwidth}
        \includegraphics[width=\textwidth]{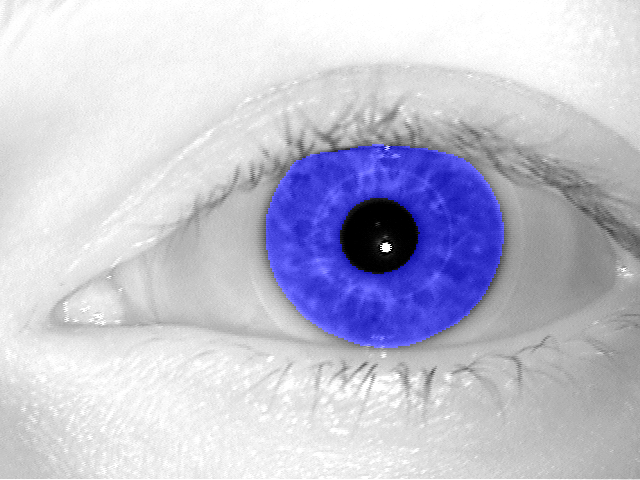}\\
        \includegraphics[width=\textwidth]{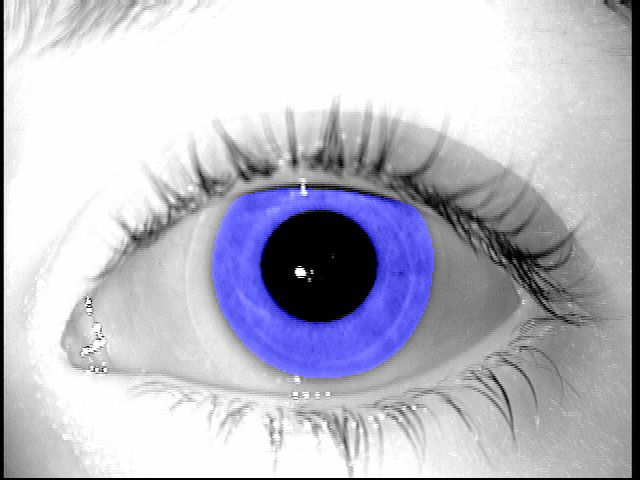}        
        \caption{}
    \end{subfigure}
    \begin{subfigure}[t]{0.19\textwidth}
        \includegraphics[width=\textwidth]{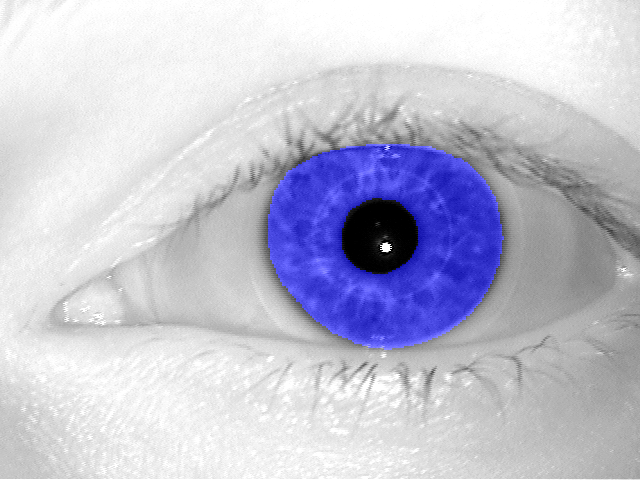}\\
        \includegraphics[width=\textwidth]{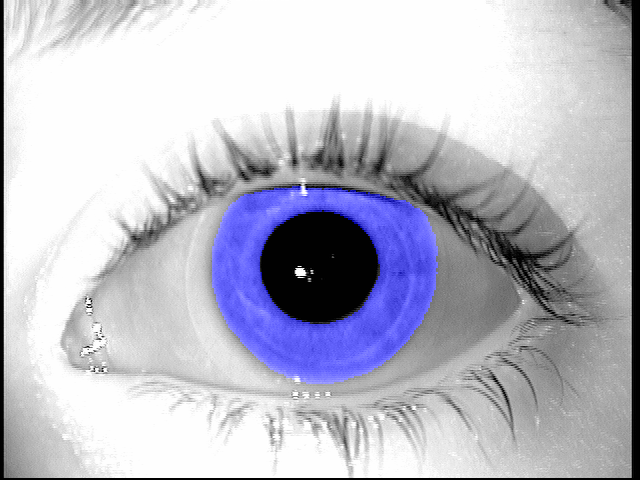}
        \caption{}
    \end{subfigure}
    \begin{subfigure}[t]{0.19\textwidth}
        \includegraphics[width=\textwidth]{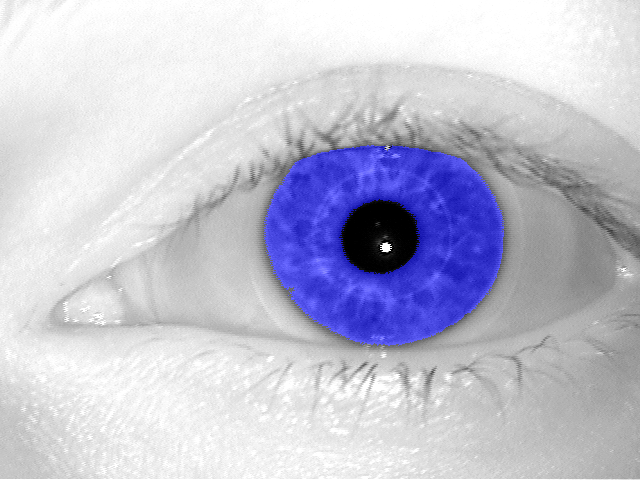}\\
        \includegraphics[width=\textwidth]{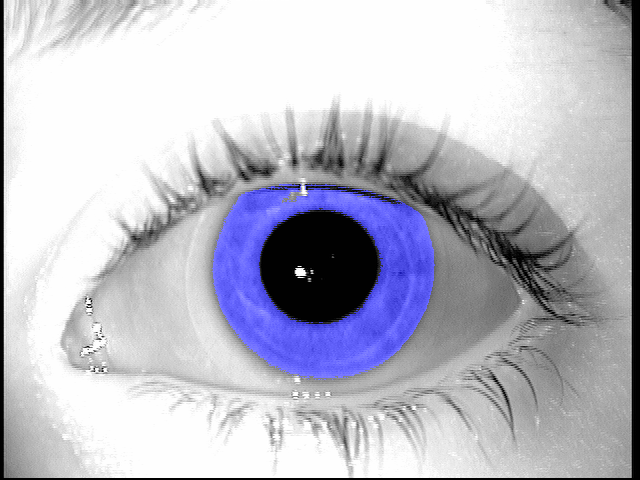}
        \caption{}
    \end{subfigure}
    \begin{subfigure}[t]{0.19\textwidth}
        \includegraphics[width=\textwidth]{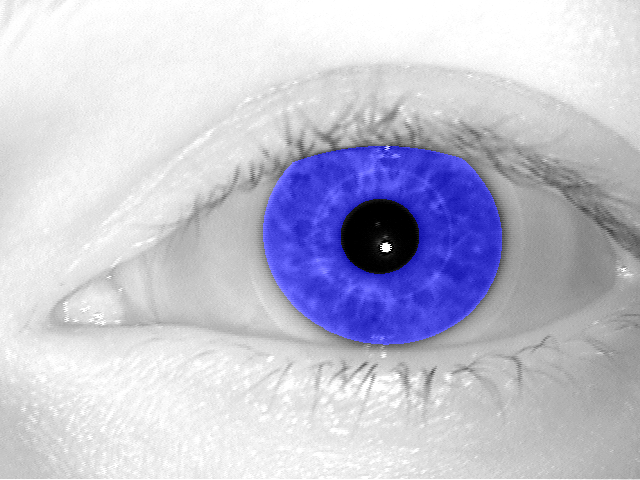}\\
        \includegraphics[width=\textwidth]{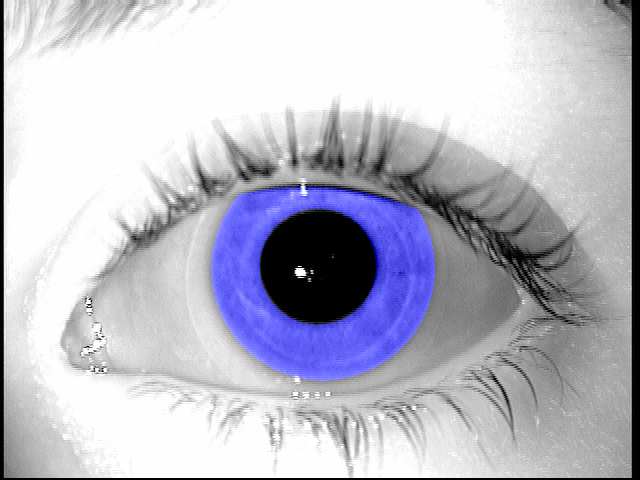}
        \caption{}
    \end{subfigure}
    \caption{Example of segmentation results for a sample from ND-Iris-0405 ({\bf upper row}) and BioSec ({\bf lower row}) benchmarks, obtained from the off-the-shelf deep learning-based methods {\bf re-trained for iris segmentation task:} {\bf (a)} modified VGG-16 with dilated convolutions \cite{dilated-cnn}, {\bf (b)} Deep Residual Network \cite{He_CVPR_2016}, {\bf (c)} SegNet \cite{badrinarayanan2015segnet}. Corresponding result of the existing learning-free segmentation tool OSIRIS \cite{osiris} {\bf (d)} and manually-annotated ground-truth \cite{halmstad1,halmstad2} {\bf (e)} are also shown.}
    \label{fig:segexamples}
\end{figure*}

\begin{abstract}
This paper offers three new, open-source, deep learning-based iris segmentation methods, and the methodology how to use irregular segmentation masks in a conventional Gabor-wavelet-based iris recognition. To train and validate the methods, we used a wide spectrum of iris images acquired by different teams and different sensors and offered publicly, including data taken from CASIA-Iris-Interval-v4, BioSec, ND-Iris-0405, UBIRIS, Warsaw-BioBase-Post-Mortem-Iris v2.0 (post-mortem iris images), and ND-TWINS-2009-2010 (iris images acquired from identical twins). This varied training data should increase the generalization capabilities of the proposed segmentation techniques. In database-disjoint training and testing, we show that deep learning-based segmentation outperforms the conventional (OSIRIS) segmentation in terms of Intersection over Union calculated between the obtained results and manually annotated ground-truth. Interestingly, the Gabor-based iris matching is not always better when deep learning-based segmentation is used, and is on par with the method employing Daugman's based segmentation.
\end{abstract}

\section{Introduction}
\label{sec:intro}

The accuracy of iris recognition, which in 2018 celebrated its 25 anniversary (assuming that Daugman's T-PAMI paper \cite{Daugman_TPAMI_1993} gave rise to this method), relies mostly on correct iris segmentation. If the iris segmentation fails, even the best feature extraction will end up with an iris code that does not correspond to actual iris texture, and consequently greatly increased chances of a false non-match. Recent advances in deep learning-based general image segmentation allowed to apply these structures also in iris segmentation. This paper addresses the deep learning-based iris segmentation by offering two novelties:

\begin{itemize}
    \item[(a)] {\bf three new, open-source iris segmentation tools} based on off-the-shelf convolutional neural networks, trained for iris segmentation task, and not open-sourced previously in iris recognition context: Deep Residual Network \cite{He_CVPR_2016}, CNN incorporating dilated convolutions \cite{dilated-cnn} and SegNet \cite{badrinarayanan2015segnet}; segmentation results of two example iris images are shown in Fig. \ref{fig:segexamples};
    \item[(b)] the methodology how to {\bf apply the resulting irregular segmentation masks to a conventional, Gabor-wavelet-based iris matching}, along with the assessment of the resulting iris recognition accuracy obtained when deep learning-based segmentation is used. This is the first, known to us, demonstration of how deep learning-based segmentation performs in the entire iris recognition pipeline, in addition to visual inspection of the results (as done in previous papers).
\end{itemize}

We compare our segmentations with results obtained by an existing open-source solution OSIRIS \cite{osiris}. We show that the offered deep learning-based methods offer a better accuracy of both the segmentation (estimated through intersection over union and Hamming distance between the predicted and ground-truth segmentations), as well as of Gabor-wavelet-based matching, when a conventional Daugman's feature extraction and matching is applied. 

The source codes of the offered segmentation methods, along with the network weights, are made available with the paper\footnote{the link to a GitHub repo has been removed to make the submission anonymous}. Training and testing were performed on publicly-available benchmarks. This makes the results presented in this work reproducible by others, and the proposed segmentation tools are easily deployed in other projects.




\section{Related Work}
\label{sec:related}

The dominant approach to iris segmentation is certainly the one based on circular approximations of the inner and outer iris boundaries \cite{daugman_1993}, with later extensions to more complex shapes approximated by Fourier series \cite{Daugman_TSMCB_2007}. We focus on more recent, deep learning-based solutions in this paper.

Jalilian and Uhl \cite{Jalilian_DLinBiometrics_2017} proposed the first, known to us, deep learning-based iris segmentation method. They used several types of convolutional encoder-decoder network trained on ND-Iris-0405, IITD and CASIA-Iris-Ageing-v5 datasets with manually annotated ground-truth segmentation masks. The authors reported better performance for CNN-based method when compared to conventional algorithms, such as OSIRIS \cite{osiris}, WAHET \cite{Uhl_ICB_2012}, CAHT \cite{Rathgeb_IB_2013} and IFPP \cite{Uhl_IAR_2012}. The paper does not mention anything about the trained network being available to others. 

Arsalan \etal \cite{Arsalan_Symmetry_2017} adapted the VGG-Face network to segmentation of visible-light iris images acquired for NICE-II benchmark and MICHE dataset. The proposed version of VGG has two output neurons, and thus the segmentation is expressed as a binary classification problem (iris / non-iris) defined for local image patches. This, certainly, results in significant processing times for each iris image. This effort was later extended to NIR iris images acquired for CASIA-Iris-Interval-v4 and IITD datasets \cite{Arsalan_Sensors_2018}. The paper does not provide an information about an open-source solution offered to ther researchers.

Severo \etal \cite{Severo_IJCNN_2018} proposed to use fine-tuned YOLO object detector \cite{Redmon_CVPR_2017} to find a bounding box that includes an eye in the NIR images. This certainly is not sufficient for iris recognition, since the exact location of non-occluded iris texture is needed for extraction of meaningful features. 

Lozej \etal \cite{Lozej_IWOBI_2018} re-trained the U-Net architecture \cite{Ronneberger_MICCAI_2015}, with different settings of hyperparametres, on CASIA benchmark and made this model publicly available to the research community. 

Bazrafkan \etal \cite{Bazrafkan_NN_2018} proposed a few newly designed convolutional neural networks, working in parallel for end-to-end iris segmentation. Initially trained on NIR images from BATH800 and CASIA-Thousand-v4, these structures were fine-tuned for visible-light images collected for UBIRIS and MobBio benchmarks. This paper additionally proposed various data augmentation techniques, specific to training neural networks specialized for iris segmentation. There is no information in the paper about the availability of the software or network weights.

Bezerra \etal \cite{Bezerra_SIBGRAPI_2018} proposed to use Generative Adversarial Networks \cite{Goodfellow_NIPS_2014} in iris segmentation for the first time, in addition to previously used fully convolutional neural networks. These solutions were evaluated on NIR images from BioSec, CASIA-Iris-Interval-v3, CASIA-Iris-Thousand-v4 and IITD datasets, as well as on visible-light images taken from NICE.I, CrEye-Iris and MICHE-I benchmarks. The authors offer manually labeled 2,431 images from ASIA-Thousand, CrEye-Iris and MICHE-I datasets, however the implemented methods and/or network weights are not offered with the paper.

Ahmad and Fuller \cite{Ahmad_ArXiv_2018} used the same supervised learning scenario to train the Mask R-CNN \cite{He_ICCV_2017} for iris segmentation. The authors used CASIA-Iris-Interval-v4, Notre Dame 0405, and UBIRIS benchmarks for which the ground-truth segmentation masks were available. The authors report a better cross-dataset accuracy when compared to previous solutions. Also in this case the paper does not include information how to get a copy of the network weights.

There are two conclusions form this summary of past work related to deep learning-based iris segmentation:

\begin{enumerate}
    \item the only previous solution that was open-sourced is the one offered by Lozej \etal \cite{Lozej_IWOBI_2018}, and this was in the form of the network weights,
    \item no previous work assessed the resulting deep learning-based segmentation from the matching perspective,  relying instead only on comparison to manually-annotated segmentation.
\end{enumerate}

Consequently, {\bf this paper offers three open-source, deep learning-based solutions not proposed earlier in the public domain}, and for the first time shows how to apply the resulting irregular masks in conventional, Gabor-wavelet-based iris matching, and thus {\bf evaluates the obtained segmentation masks from the perspective of the entire iris recognition pipeline}, not only visual inspection, as done in earlier papers.

\section{Selected Convolutional Neural Networks}
\label{sec:networks}

For the purpose of this study, four different convolutional architectures designed for semantic segmentation were selected for re-training and evaluation. Two of them, employing dilated convolutions, were proposed by Yu and Koltun \cite{dilated-cnn}. These two networks are referred to as the {\em front-end} module and the {\it context} module. The front-end module is a modified VGG-16 pyramid-like network. The most significant change the authors made was removing the final two pooling and striding layers. Convolutional layers were then dilated by a factor of 2 for each removed pooling layer they appear after. The final features of the front-end module serve as input to the context module, which has seven dilated convolutional layers with kernels of size $3\times3$, followed by an output layer with $1\times1$ kernels. The dilation factors for the first seven layers are 1, 1, 2, 4, 8, 16, and 1, respectively. For our experiments with the front-end and context modules, we used a fork of Caffe provided by Yu and Koltun\footnote{\scriptsize\url{https://github.com/fyu/dilation}}.
The third network we evaluated was a dilated residual network (DRN) presented by Yu, Koltun, and Funkhouser \cite{drn}. A DRN is a modified ResNet, where two striding layers were removed. Convolutional layers after the first removed striding layer were modified to have a dilation of 2. Convolutional layers after both removed striding layers were made to have a dilation of 4. This decreases the amount of downsampling without increasing the number of parameters. The network is further modified to remove gridding artifacts caused by the dilation. These modifications include removing max pooling, adding convolutional layers at the end, and removing residual connections from the last few layers. For the experiments, we used a PyTorch implementation provided by Yu \etal, the DRN-D-22 model in particular\footnote{\scriptsize\url{https://github.com/fyu/drn}}. In the course of experiments, we found that 100,000 iterations were sufficient to get the well-performing models, and that context module was better than front-end module. This is why we further skip the front-end-based model in testing.

The fourth and final network is SegNet, which contains an encoder and decoder network \cite{badrinarayanan2015segnet}. The encoder network is a modified VGG-16 network where the fully-connected layers were removed. The decoder network contains convolutional and upsampling layers. Indices from the max-pooling layers in the encoder network are stored and used in the up-sampling layers in the decoder network. For training and testing of SegNet, the implementation provided in the MATLAB Neural Network Toolbox was used.

\section{Iris Image Databases}
\label{sec:databases}

Data for training and evaluation comes from several publicly available databases of iris images, and was divided into {\bf subject-disjoint training and testing subsets}.

The {\bf training} data comes from three databases: 2639 images are from the CASIA-Iris-Interval-v4 database\footnote{\scriptsize\url{http://www.cbsr.ia.ac.cn/english/IrisDatabase.asp}}, 898 images are from the University of ND-Iris-0405 database\footnote{\scriptsize\url{https://cvrl.nd.edu/projects/data/}}, and 800 images are mixed from Warsaw-BioBase-Post-Mortem-Iris v2.0\footnote{\scriptsize\url{http://zbum.ia.pw.edu.pl/EN/node/46}} or ND-TWINS-2009-2010 database\footnote{\scriptsize\url{https://cvrl.nd.edu/projects/data/}}.


Data from three databases are used for {\bf testing}: 385 images are from the ND-Iris-0405 database (these images are person-disjoint with the images used in the training set), 1,200 images are from the BioSec baseline corpus database \cite{biosec}, and 2,250 images are from the UBIRIS.v2 database \cite{ubiris}. The UBIRIS database is the only database used for testing that consists of visible wavelength images. During testing, only the red channel of the RGB color space from the UBIRIS images were used.

All images have their corresponding ground truth masks, denoting the iris location within the image, as well as occlusions, if such are present. Ground truth for the BioSec database comes from the IRISSEG-CC dataset by Halmstad University, whereas ground truth for the CASIA-Iris-Interval-v4, ND-Iris-0405, and UBIRIS databases are from the IRISSEG-EP dataset by the University of Salzburg \cite{halmstad1, halmstad2}.

\section{Evaluation Strategies}
\label{sec:evaluationMethodology}
Evaluation of the proposed approach consists of two methodologies. 
In the first, we only assess the accuracy of the segmentation predictions produced by our models.
In the second, the segmentation results are injected into a typical iris recognition pipeline, whose recognition accuracy is then evaluated in comparison to the unmodified, open-source iris recognition method (OSIRIS). 

\subsection{Segmentation accuracy}
For the segmentation accuracy evaluation, we compare the predictions obtained from each of the models trained in this study against manually annotated, ground truth binary masks, described earlier in Sec. \ref{sec:databases}. For each pair of a predicted segmentaion and a ground truth mask, two accuracy metrics typically employed for accuracy assessment in segmentation tasks are used:
\begin{itemize}
    \item {\bf Intersection over Union (IoU)}, which denotes the proportion of the intersection of the ground truth and the prediction to the union of the two,
    \item {\bf fractional Hamming Distance (HD)}, which denotes the proportion of disagreeing pixels to the overall number of pixels.
\end{itemize}

For each test database and each model employed for generating segmentation results, average values of the above metrics, together with standard deviation $\sigma$ values are reported. 

\subsection{Matching accuracy}
\begin{figure}[t]
    \centering
    Easy sample:\\
        \includegraphics[width=0.115\textwidth]{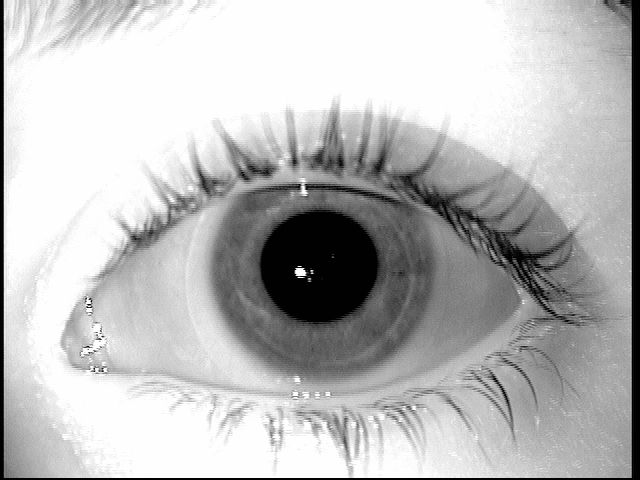}\hskip1mm
        \includegraphics[width=0.115\textwidth]{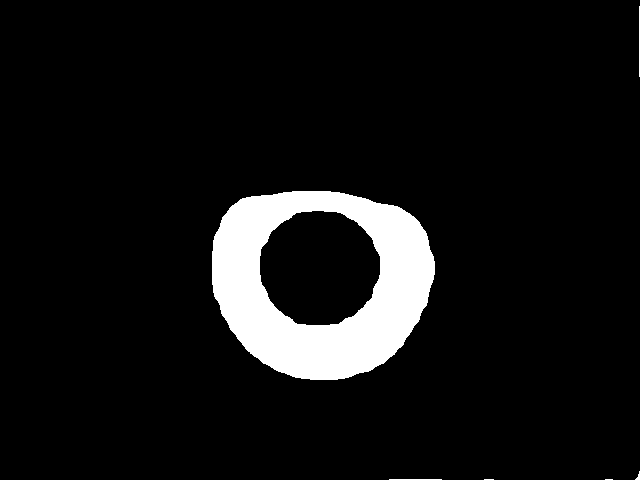}\hskip1mm
        \includegraphics[width=0.115\textwidth]{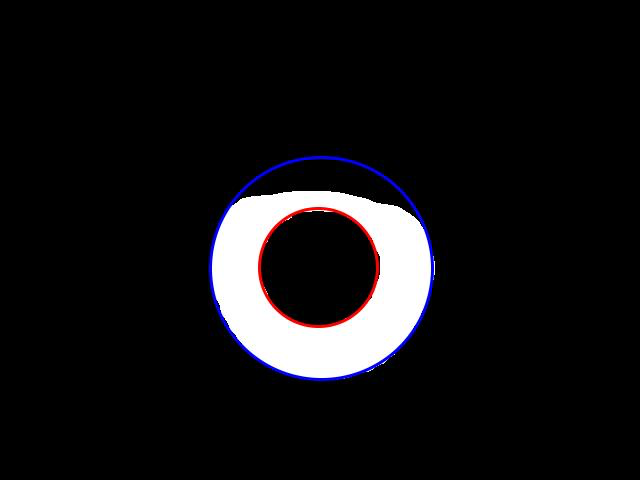}\hskip1mm
        \includegraphics[width=0.115\textwidth]{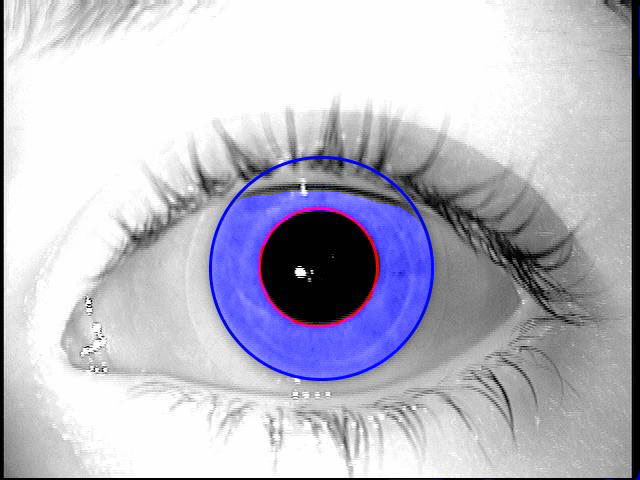}\\\vskip1mm
        Difficult sample:\\
        \includegraphics[width=0.115\textwidth]{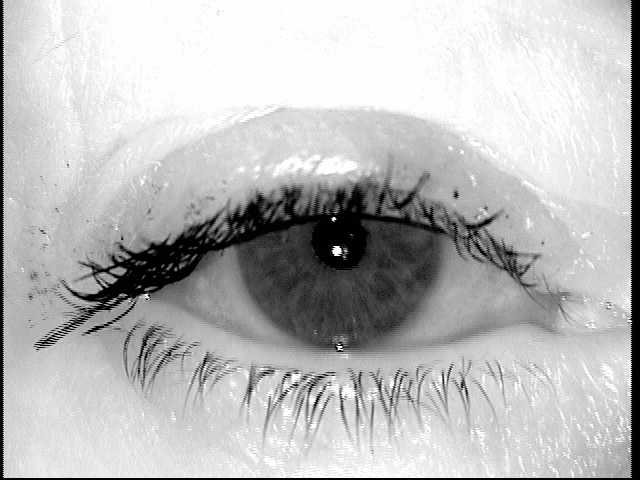}\hskip1mm
        \includegraphics[width=0.115\textwidth]{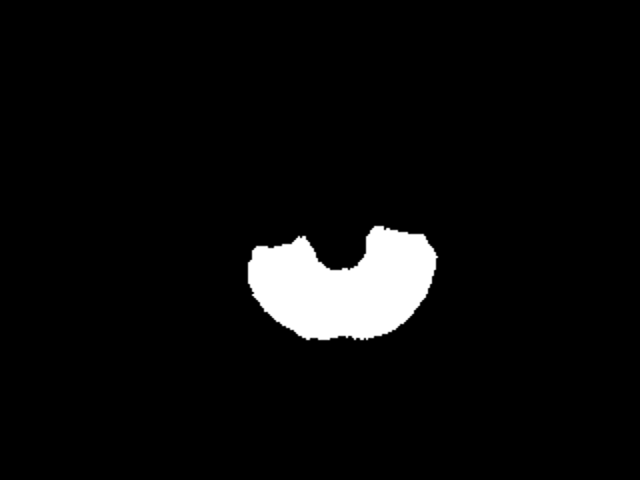}\hskip1mm
        \includegraphics[width=0.115\textwidth]{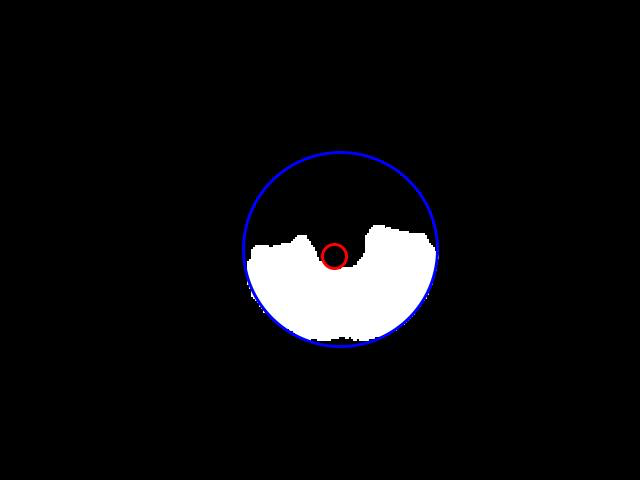}\hskip1mm
        \includegraphics[width=0.115\textwidth]{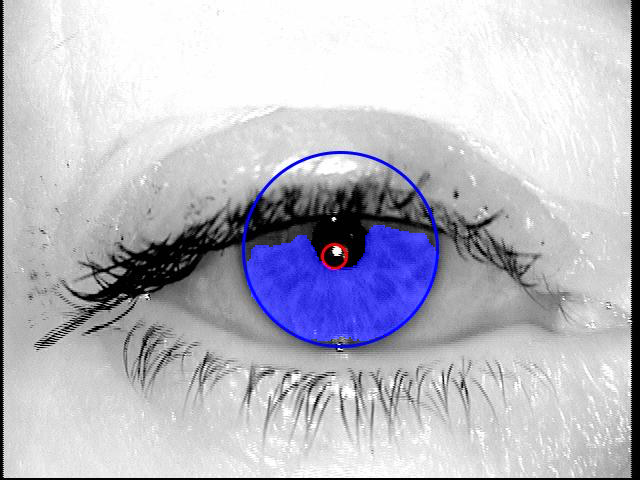}\\
    \caption{Iris boundary fitting using circular Hough transform applied to the binary predictions obtained from each trained neural network.}
    \label{fig:houghing}
\end{figure}

For evaluation of the matching performance that the proposed segmentation methods can yield, the results obtained in the first part of the evaluation, namely binary iris masks for test images, are injected into the OSIRIS iris recognition algorithm \cite{osiris}. OSIRIS, which was created in the framework of the BioSecure project, follows the traditional Daugman's approach to iris recognition, with iris image normalization onto a polar-coordinate rectangle, and subsequent encoding with Gabor-based filtering for three different complex Gabor kernels. The comparison results are given in the form of a fractional Hamming distance between the two binary iris codes. Since the OSIRIS is an open-source project, it enables easy modifications at different stages of the recognition pipeline, such as injecting custom segmentation results, that are later used for iris normalization and masking. For normalization of the images and their respective masks, iris and pupil circle parameters are required by the OSIRIS method. To meet this requirement, we employ the circular Hough transform to approximate the inner and outer iris boundary in the binary mask and in the image, cf. Fig. \ref{fig:houghing}.


\section{Procedure for training and evaluation}
\label{sec:trainingAndTestingProcedures}

\subsection{Data Augmentation}
\label{sec:augmentation}
\begin{figure}[t]
    \centering
    Original (left) and after edge enhancement:\\
        \includegraphics[width=0.155\textwidth]{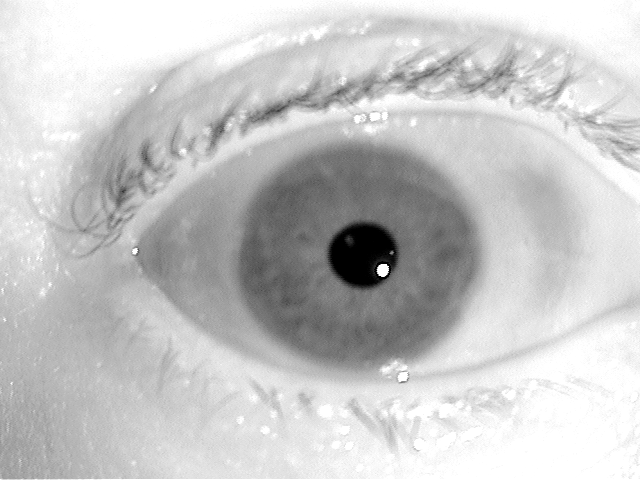}\hskip1mm
        \includegraphics[width=0.155\textwidth]{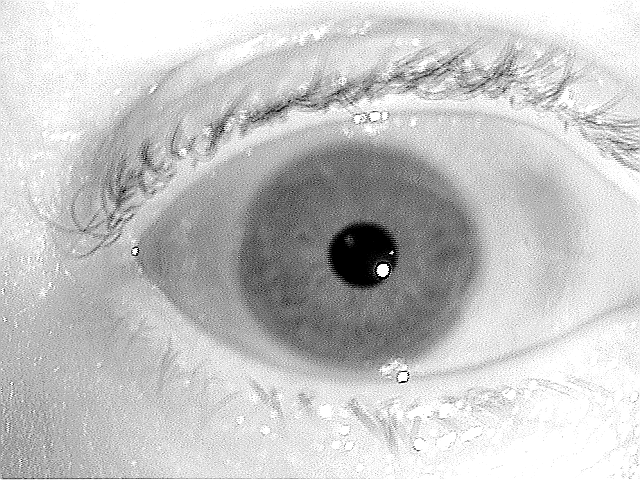}\hskip1mm
        \includegraphics[width=0.155\textwidth]{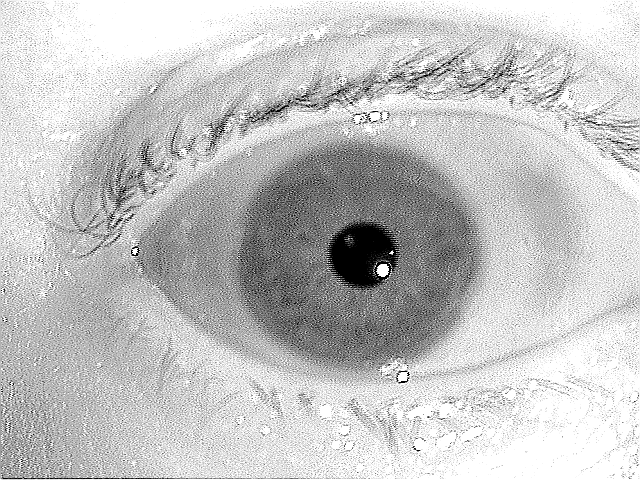}\\\vskip1mm
        Gaussian blurring:\\
        \includegraphics[width=0.155\textwidth]{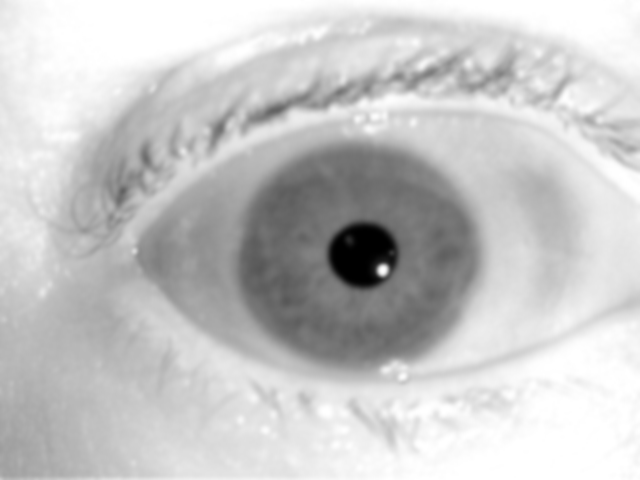}\hskip1mm
        \includegraphics[width=0.155\textwidth]{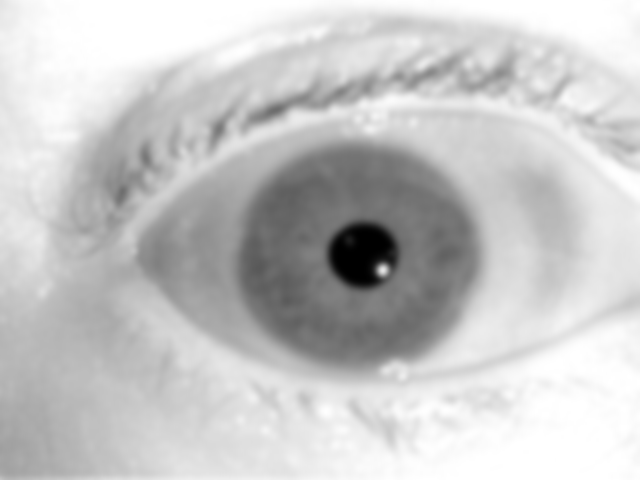}\hskip1mm
        \includegraphics[width=0.155\textwidth]{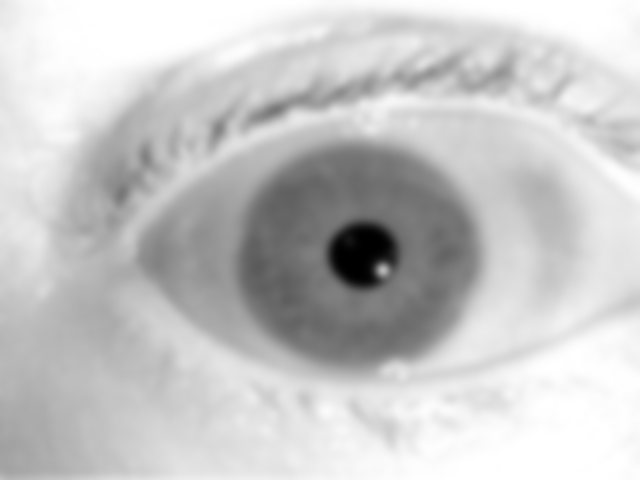}\\
    \caption{Illustration of the data augmentation transformations applied for training samples.}
    \label{fig:augmentation}
\end{figure}

For the training stage, each image in the training dataset was augmented five-fold, leading to a total of 26,022 training images. The augmentations were performed using the Pillow library for Python \cite{pillow}. Three of the modifications add Gaussian blur with a radius of 2, 3, and 4, whereas the remaining two augmentations add different edge enhancement, Fig. \ref{fig:augmentation}.

\subsection{Training and Validation}
\label{sec:trainAndVal}

For the training stage, 80\% of the training data were used for the actual training, whereas 20\% were set aside for validation. We used images of size $640\times480$ to train the front-end module and the context module of the dilated convolutions model, whereas images of size $320\times240$ were used to train SegNet and DRN. Each network was initialized with the provided, pre-trained weights. Initially, these networks were pre-trained on non-iris databases, such as ImageNet, which contains natural images of various objects \cite{imagenet}. We used stochastic gradient descent as the minimization method for the training of each network. 

We trained the front-end and context modules for 100k iterations, or about 5 epochs, with a batch size of 1 due to memory constraints. The front-end module had a learning rate of $1e^{-4}$ and the context module had a learning rate of $1e^{-3}$. Both modules were trained with a momentum of 0.9. SegNet was trained for 20 epochs with a batch size of 4, a learning rate of $1e^{-3}$, and a momentum of 0.9. DRN was trained for 100 epochs with a batch size of 8, a learning rate of $1e^{-2}$, and a momentum of 0.9.

\subsection{Testing}
\label{sec:testing}
During testing, a binary prediction was obtained from each of the three networks and from the default OSIRIS algorithm for every image belonging to the test dataset, cf. Sec. \ref{sec:databases}. Then, the accuracy metrics introduced in Sec. \ref{sec:evaluationMethodology}, namely the IoU and HD, were calculated to assess the coherence between the predicted and the ground truth masks. Predictions obtained from the front-end and context modules were not resized in any way, whereas the predictions obtained from DRN and from SegNet were upscaled from $320\times240$ to $640\times480$ pixels.

\section{Experimental results}
\label{sec:results}

\begin{table*}[t]
\renewcommand{\arraystretch}{1.4}
\caption{Mean IoU and HD metrics obtained for each test dataset and each method, together with their standard deviations $\sigma$. The best and the worst IoU results for each dataset are colored in green and red, respectively.}
\label{table:segmentationAccuracies}
\centering
\begin{tabular}[t]{|c|c|c|c|c|c|}
\hline
\textbf{Metric [\%]} & \textbf{Dataset} & \textbf{OSIRIS (baseline)} & \textbf{DRN} & \textbf{Context-100k} & \textbf{SegNet}\\
\hline
\hline
\multirow{3}{*}{IoU} & \textbf{BioSec} & \cellred{\bf 84.81 $\pm$ 6.73} & \cellgreen{\bf 87.29 $\pm$ 5.79} & 85.92 $\pm$ 5.91 & 85.93 $\pm$ 6.79 \\
\cline{2-6}
& \textbf{ND-Iris-0405} & \cellred{\bf 86.28 $\pm$ 6.5} & 89.61 $\pm$ 5.08 & 89.45 $\pm$ 3.85 & \cellgreen{\bf 89.75 $\pm$ 4.95}\\
\cline{2-6}
& \textbf{UBIRIS} & \cellred{\bf 42.69 $\pm$ 36.79} & 44.4 $\pm$ 29.75 & \cellgreen{\bf 56.28 $\pm$ 26.93} & 48.27 $\pm$ 26.42 \\
\hline
\hline
\multirow{3}{*}{HD} & \textbf{BioSec} & 0.80 $\pm$ 1.39 & 1.12 $\pm$ 0.48 & 1.24 $\pm$ 0.53 & 1.22 $\pm$ 0.55\\
\cline{2-6}
& \textbf{ND-Iris-0405} & 1.29 $\pm$ 0.64 & 0.93 $\pm$ 0.43 & 0.97 $\pm$ 0.37 & 0.91 $\pm$ 0.39 \\
\cline{2-6}
& \textbf{UBIRIS} & 8.83 $\pm$ 11.06 & 3.35 $\pm$ 1.72 & 3.12 $\pm$ 2.37 & 3.25 $\pm$ 1.87 \\
\hline
\end{tabular}
\end{table*}

\subsection{Segmentation results}
Table \ref{table:segmentationAccuracies} aggregates the IoU and HD metrics averaged over all samples from each of the test databases. Notably, OSIRIS yielded the worst average results for all databases, whereas the best results are split between the three deep learning-based models. For the Biosec database, the best was DRN, with $IoU=87.29\%$, for the ND-Iris-0405 database it's the SegNet model with $IoU=89.75\%$, and finally for the most challenging UBIRIS database the best results were obtained from the Context-100k model, which gave $IoU=56.28\%$. The performance observed for a deep learning-based model run on the UBIRIS was also the largest IoU-wise improvement, as OSIRIS was able to only produce $IoU=42.69\%$ for this set.

\begin{figure}[t]
    \centering
        \includegraphics[width=0.495\textwidth]{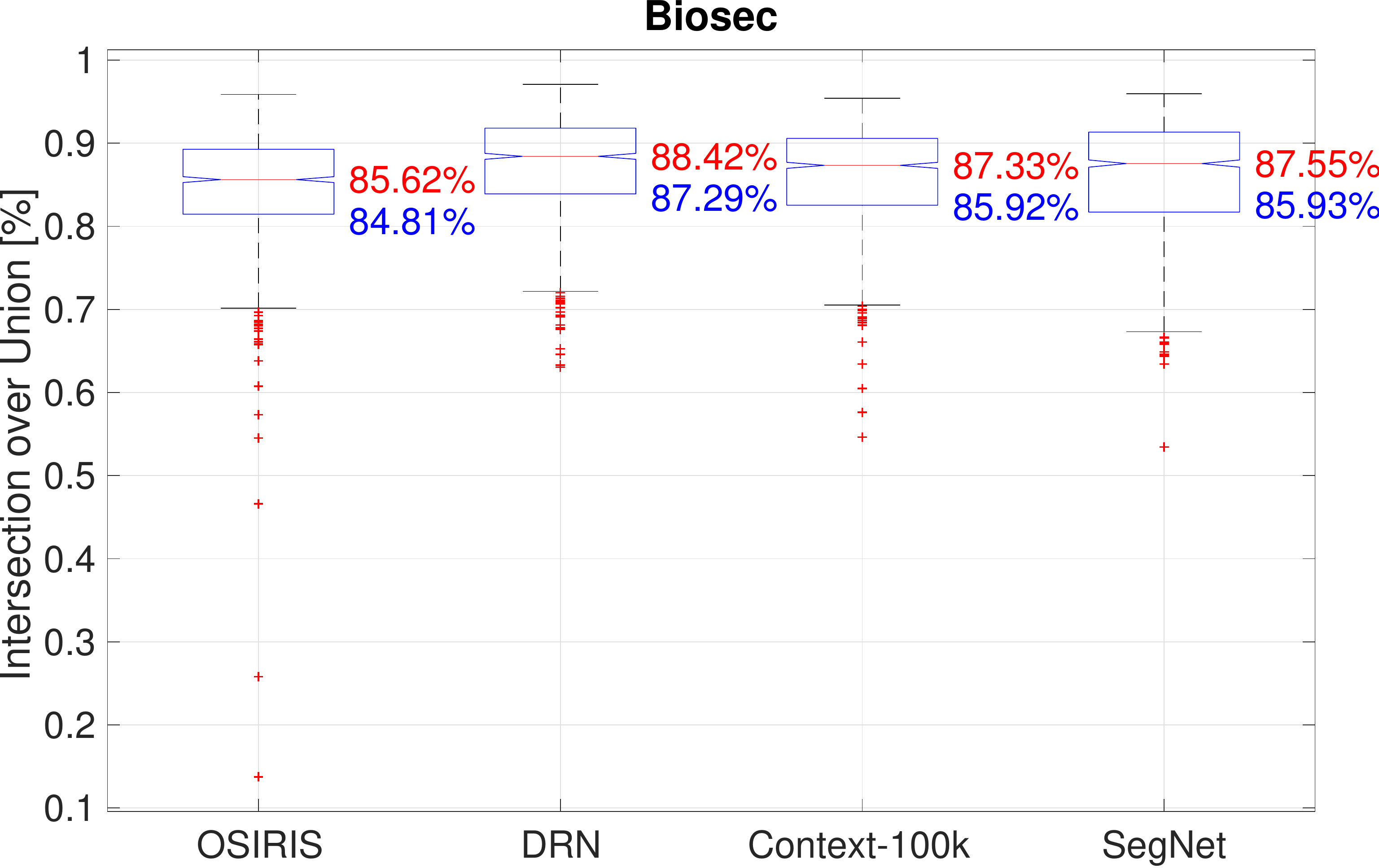}\\\vskip5mm
        \includegraphics[width=0.495\textwidth]{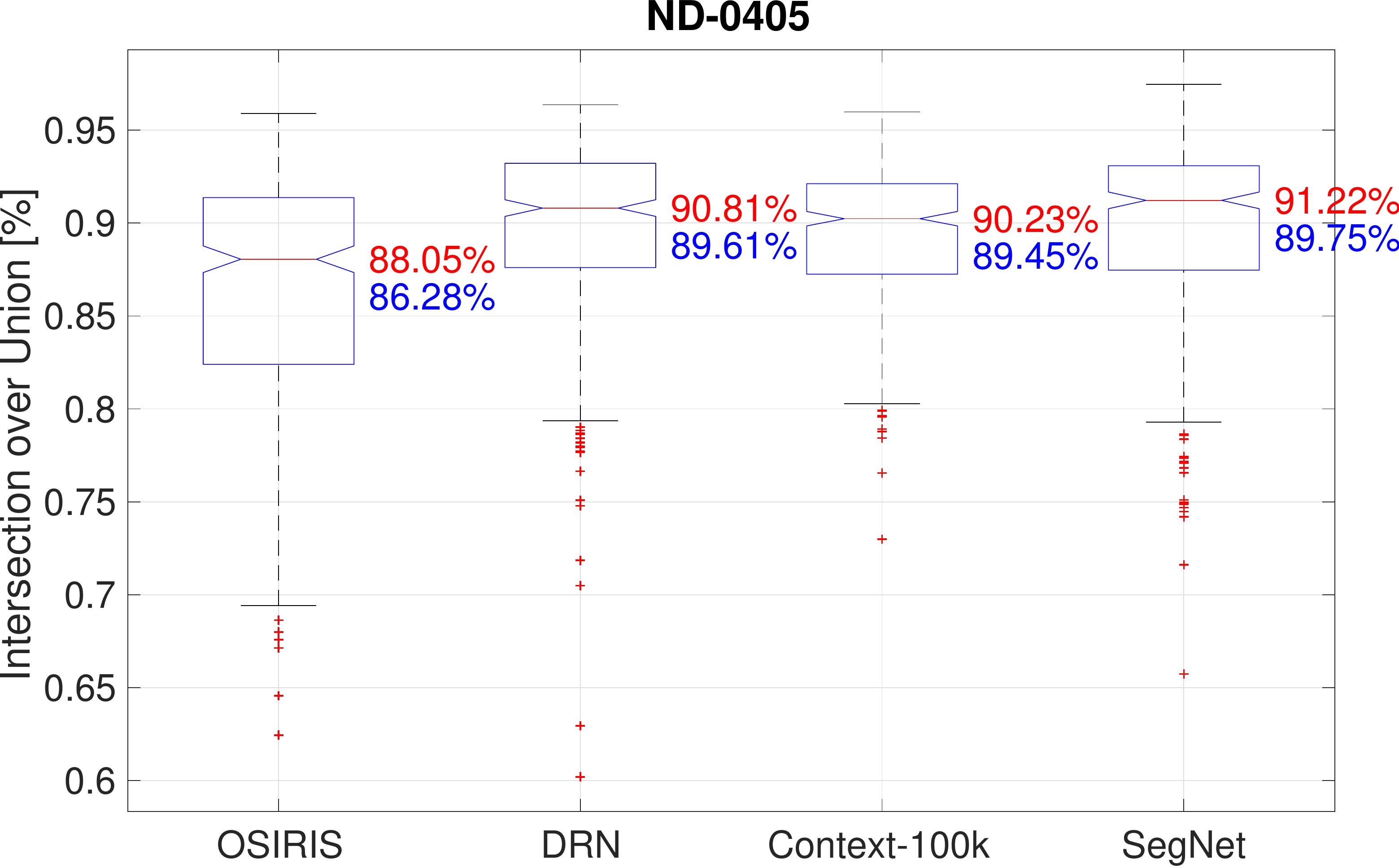}\\\vskip5mm
        \includegraphics[width=0.495\textwidth]{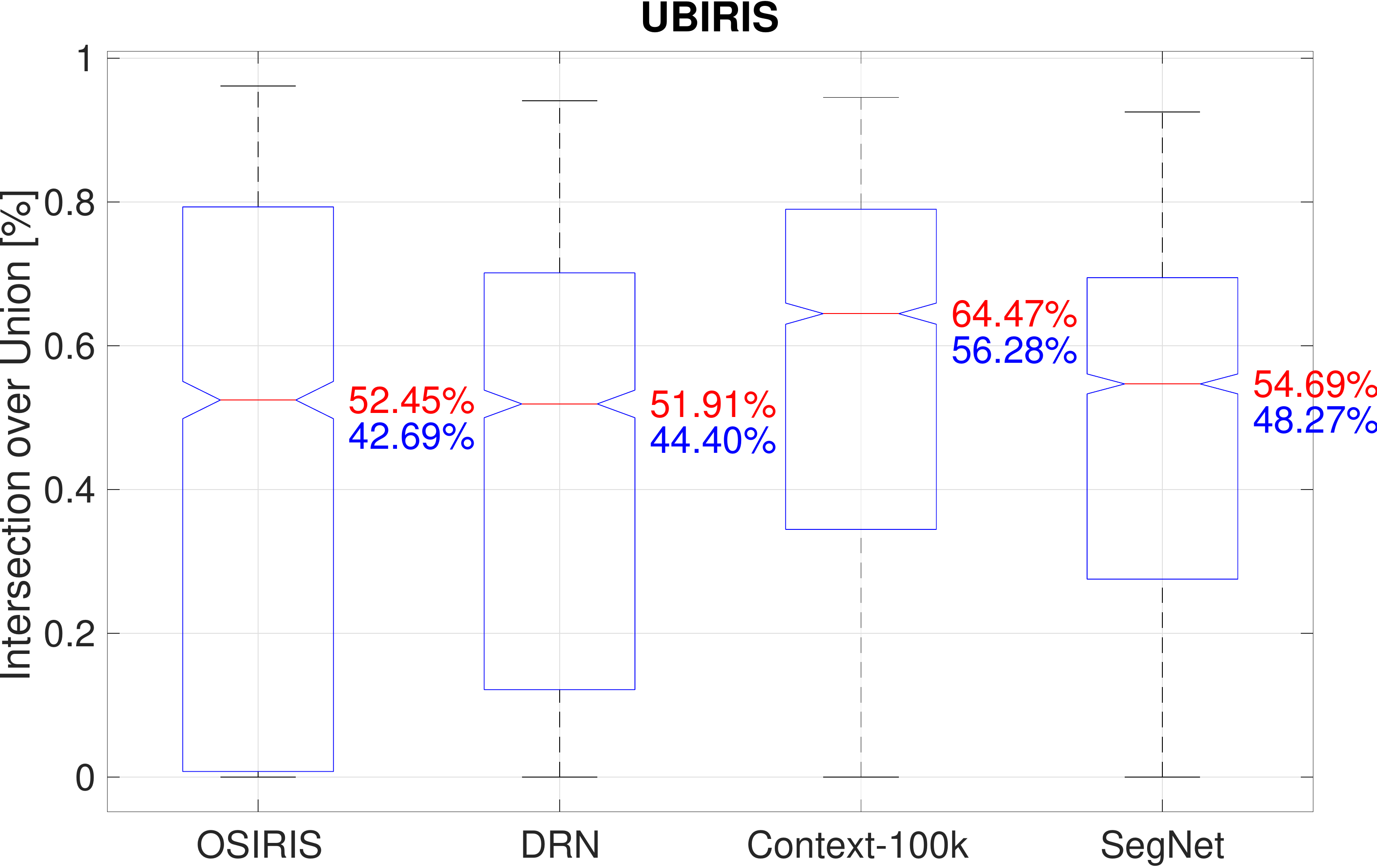}\\
    \caption{Boxplots representing Intersection over Union for three test databases and four methods employed in this study. Median values are shown in red, height of each box corresponds to an inter-quartile range (IQR) spanning from the first (Q1) to the third (Q3) quartile, whiskers span from Q1-1.5*IQR to Q3+1.5*IQR, and outliers are shown as red crosses. Notches represent 95\% confidence intervals of the median.}
    \label{fig:boxplots}
\end{figure}

\subsection{Gabor-based matching performance}
For the matching performance assessment stage, we only use the results obtained for the ND-Iris-0405 and the Biosec databases, since the UBIRIS database contains images that are not compliant with the ISO/IEC 29794-6:2015 standard on iris image quality, and thus are not suitable to be used with a traditional iris recognition algorithm. 

\begin{figure}[t]
    \centering
        \includegraphics[width=0.495\textwidth]{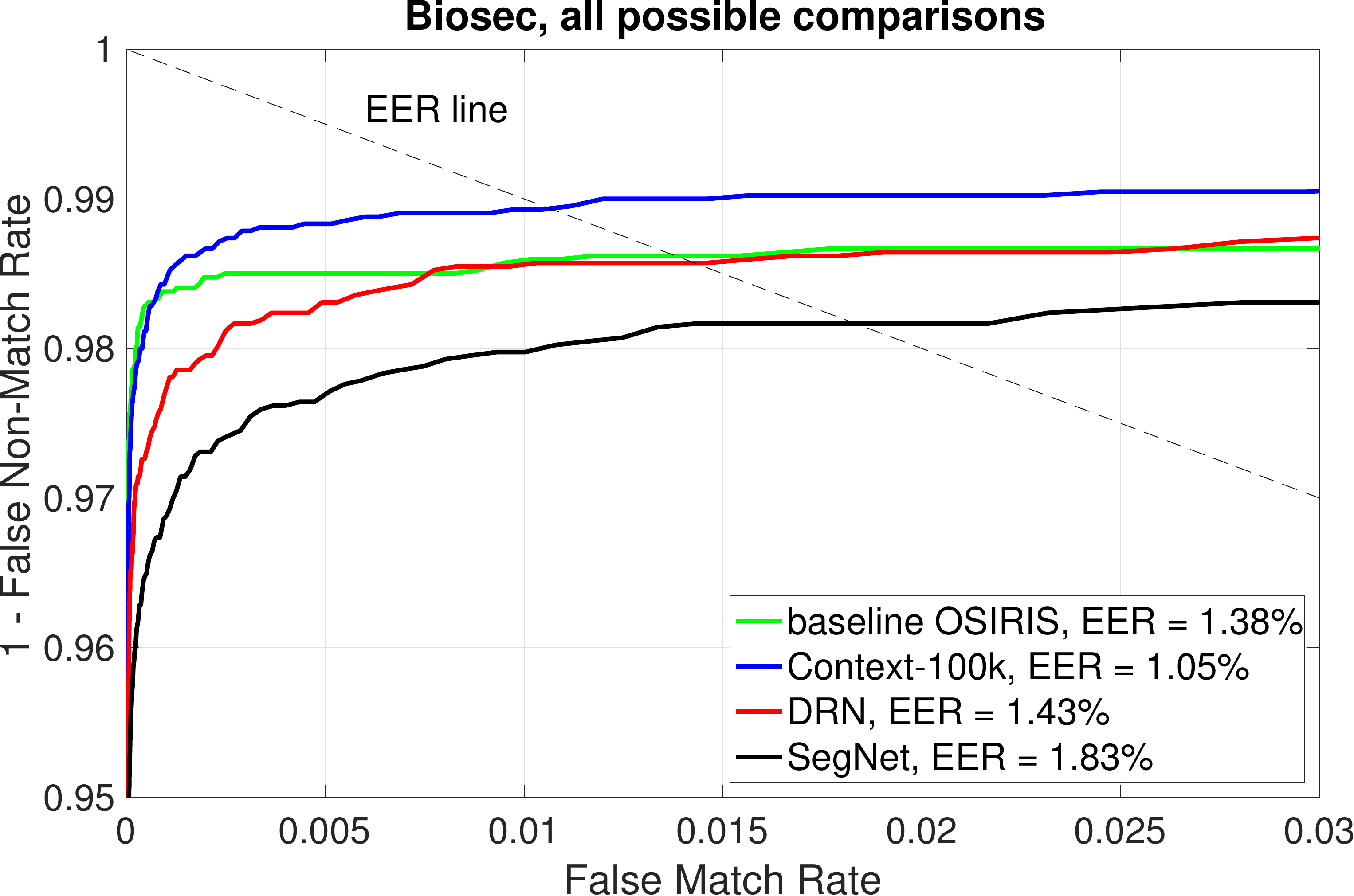}\\
    \caption{ROC curves plotted for all possible comparisons generated with samples from the Biosec database, for all deep learning-based segmentation methods discussed in this work, as well as for the baseline OSIRIS segmentation results.}
    \label{fig:roc_biosec}
\end{figure}

\begin{figure}[t]
    \centering
        \includegraphics[width=0.495\textwidth]{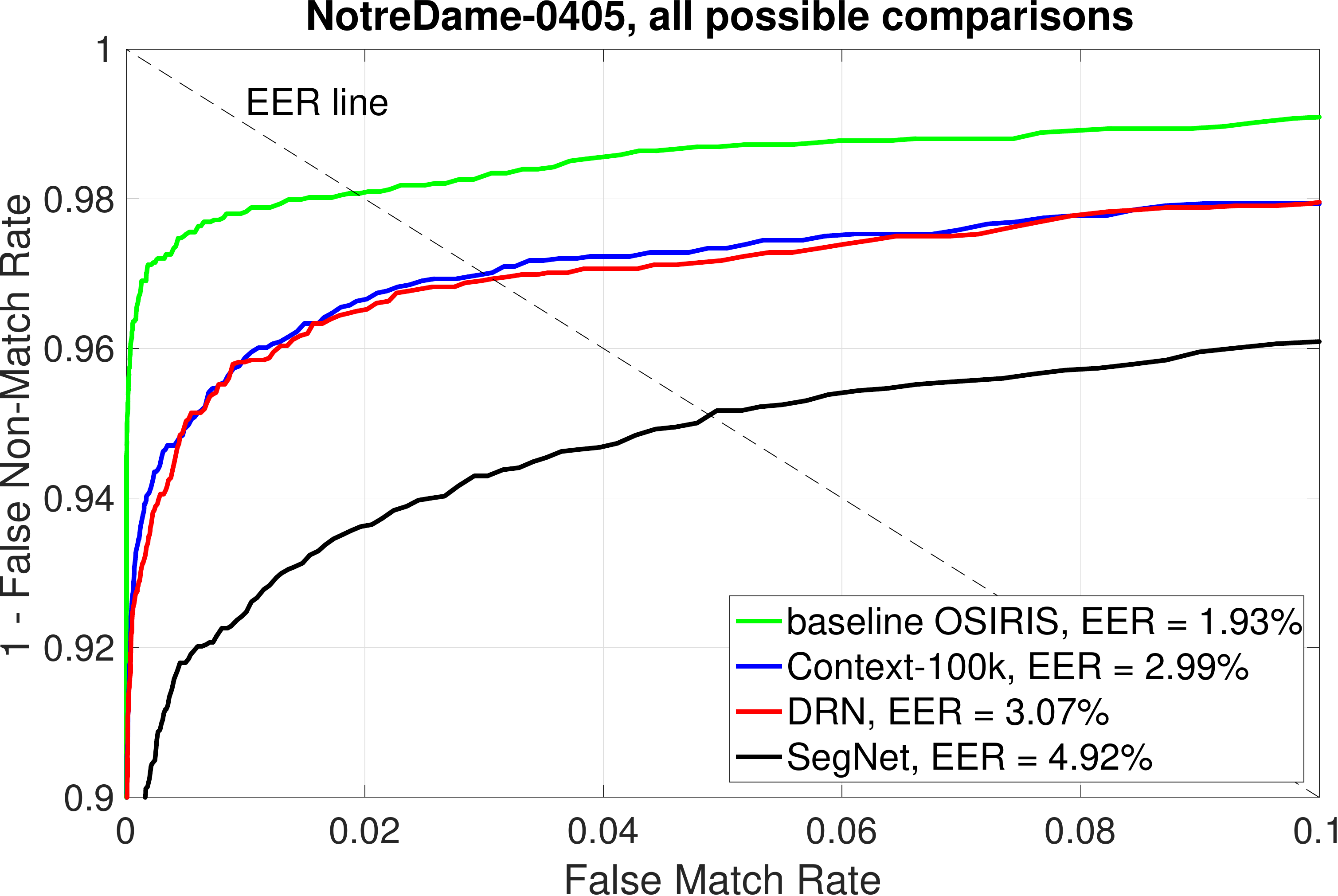}\\
    \caption{Same as in \ref{fig:roc_biosec}, but for comparisons of samples from the ND-Iris-0405 database.}
    \label{fig:roc_nd}
\end{figure}

Figures \ref{fig:roc_biosec} and \ref{fig:roc_nd} present Receiver Operating Characteristic (ROC) curves obtained from all possible comparison scores in the Biosec and in the ND-Iris-0405 databases, respectively. For the Biosec database, which contains a much larger number of samples than the ND database (1200 vs 385), all segmentation methods are performing similarly, with the Context-100k showing a minor advantage over the stock OSIRIS with EER=1.05\% vs 1.38\%, whereas the two other deep learning-based methods are producing results that are slightly worse than OSIRIS, with EER=1.43\% and 1.83\% for DRN and SegNet, respectively.

The ND-Iris-0405 database, on the other hand, contains only 385 samples in the test subset, and here OSIRIS offers the best performance, with EER=1.93\%, compared to EER=2.99\% obtained by the best-performing Context-100k model. However, a small number of samples can cause the errors to be over-estimated by only a few samples for which the deep learning-based models returned bad segmentation outputs. 

\section{Conclusions}
\label{sec:conclusions}

This paper discusses, to our knowledge for the first time, the possibility of using irregular iris segmentation masks returned by deep learning-based models in a conventional Gabor-based iris recognition. We trained three different deep learning-based structures (modified VGG with dilated convolutions, DRN and SegNet) with a wide spectrum of NIR iris images, and showed that (a) these models outperform conventional iris recgmentation based on circular approximations of iris boundaries, and (b) these irregular masks can still be applied in the most popular Gabor-based iris recognition pipeline. The source codes of the new proposed iris segmentation tools, and the network weights are offered along with this paper, which increases a set of publicly-available iris segmentation tools.

\end{document}